\documentclass[sigconf, nonacm]{acmart}

\newcommand\vldbdoi{XX.XX/XXX.XX}
\newcommand\vldbpages{XXX-XXX}
\newcommand\vldbvolume{18}
\newcommand\vldbissue{1}
\newcommand\vldbyear{2025}
\newcommand\vldbauthors{\authors}
\newcommand\vldbtitle{\shorttitle} 
\newcommand\vldbavailabilityurl{URL_TO_YOUR_ARTIFACTS}
\newcommand\vldbpagestyle{plain} 

\usepackage{microtype}
\usepackage{amsfonts}
\usepackage{amsmath}
\usepackage{booktabs}
\usepackage{multirow}
\usepackage{calligra}
\usepackage{pdfpages}

\usepackage{graphicx}
\usepackage{booktabs}
\usepackage{subcaption}
\usepackage{colortbl}
\usepackage{balance}
\usepackage{caption}

\usepackage{listings}
\usepackage{epstopdf}
\usepackage{iitem}
\usepackage{framed}
\usepackage{setspace}
\usepackage{verbatim}
\usepackage{enumitem}

\usepackage{amstext}
\usepackage[misc]{ifsym}

\usepackage{xcolor}
\usepackage{hyperref}
\usepackage[linesnumbered,ruled,noend]{algorithm2e}
\usepackage{algpseudocode}

\usepackage[capitalize]{cleveref}
\crefname{section}{Sec.}{Secs.}
\Crefname{section}{Section}{Sections}
\Crefname{table}{Table}{Tables}
\crefname{table}{Tab.}{Tabs.}

\newcommand{\red}[1]{\textcolor{red}{#1}}

\usepackage{CJKutf8}
\newcommand{\zwt}[1]{\textcolor{red}{[\begin{CJK*}{UTF8}{gkai}#1\end{CJK*}---zwt.]}}

\begin{document}
\title{DataSculpt: Crafting Data Landscapes for Long-Context LLMs through Multi-Objective Partitioning}

\author{Keer Lu$^{*1}$, Xiaonan Nie$^{*1}$, Zheng Liang$^2$, Da Pan$^2$, Shusen Zhang$^2$, Keshi Zhao$^1$, Weipeng Chen$^2$, Zenan Zhou$^2$, Guosheng Dong$^{\dagger2}$, Bin Cui$^{\dagger1}$, Wentao Zhang$^{\dagger1}$}
\affiliation{
  \institution{$^1$Peking University $^2$Baichuan Inc.}
}
\email{{keer.lu, zhaoks}@stu.pku.edu.cn, {xiaonan.nie, bin.cui, wentao.zhang}@pku.edu.cn}
\email{{liangzheng, panda, zhangshusen, chenweipeng, zhouzenan, dongguosheng}@baichuan-inc.com}

\renewcommand{\authors}{Keer Lu, Xiaonan Nie, Zheng Liang, Da Pan, Shusen Zhang, Keshi Zhao, Weipeng Chen, Zenan Zhou, Guosheng Dong, Bin Cui, Wentao Zhang}











\begin{abstract}

In recent years, foundation models, particularly large language models (LLMs), have demonstrated significant improvements across a variety of tasks. One of their most important features is long-context capability, which enables them to process and generate extended text while maintaining coherence, retrieving relevant information, and efficiently handling tasks with substantial amounts of text. The key to improving long-context performance lies in effective data organization and management strategies that integrate data from multiple domains and optimize the context window during training.
Through extensive experimental analysis, we identified three key challenges in designing effective data management strategies that enable the model to achieve long-context capability without sacrificing performance in other tasks: (1) a shortage of long documents across multiple domains, (2) effective construction of context windows, and (3) efficient organization of large-scale datasets.

To address these challenges, we introduce \textbf{\textit{DataSculpt}}, a novel data management framework designed for long-context training.
We first formulate the organization of training data as a multi-objective combinatorial optimization problem, focusing on attributes including relevance, homogeneity, integrity, and efficiency. Specifically, our approach utilizes a coarse-to-fine methodology to optimize training data organization both efficiently and effectively. We begin by clustering the data based on semantic similarity (coarse), followed by a multi-objective greedy search within each cluster to score and concatenate documents into various context windows (fine).
Our comprehensive evaluations demonstrate that \textit{DataSculpt} significantly enhances long-context training performance, resulting in improvements of $18.09\%$ in retrieval augmentation, $21.23\%$ in summarization, $21.27\%$ in reading comprehension, and a $3.81\%$ increase in code completion, while also maintaining overall model proficiency with a $4.88\%$ improvement.

\end{abstract}

\maketitle

\pagestyle{\vldbpagestyle}
\begingroup\small\noindent\raggedright\textbf{PVLDB Reference Format:}\\
\vldbauthors. \vldbtitle. PVLDB, \vldbvolume(\vldbissue): \vldbpages, \vldbyear.\\
\href{https://doi.org/\vldbdoi}{doi:\vldbdoi}
\endgroup
\begingroup
\renewcommand\thefootnote{}\footnote{\noindent
$^*$ Equal Contribution. \\
$^\dagger$ Corresponding Author. \\
This work is licensed under the Creative Commons BY-NC-ND 4.0 International License. Visit \url{https://creativecommons.org/licenses/by-nc-nd/4.0/} to view a copy of this license. For any use beyond those covered by this license, obtain permission by emailing \href{mailto:info@vldb.org}{info@vldb.org}. Copyright is held by the owner/author(s). Publication rights licensed to the VLDB Endowment. \\
\raggedright Proceedings of the VLDB Endowment, Vol. \vldbvolume, No. \vldbissue\ %
ISSN 2150-8097. \\
\href{https://doi.org/\vldbdoi}{doi:\vldbdoi} \\
}\addtocounter{footnote}{-1}\endgroup

\ifdefempty{\vldbavailabilityurl}{}{
\vspace{.3cm}
\begingroup\small\noindent\raggedright\textbf{PVLDB Artifact Availability:}\\
The source code, data, and/or other artifacts have been made available at \url{https://github.com/8023looker/DataSculpt}.
\endgroup
}

\section{Introduction}

In recent years, artificial intelligence has achieved remarkable milestones across a wide range of fields~\cite{dwivedi2021artificial,lewkowycz2022solving}, including computer vision~\cite{dosovitskiy2020image,liu2021swin,riquelme2021scaling}, natural language processing~\cite{brown2020language,devlin2018bert,radford2019language,raffel2020exploring} and generative AI~\cite{achiam2023gpt}.
These breakthroughs are the result of significant progress in the development of AI systems, which is built upon three critical components: data, algorithms, and computational resources. 
\textit{Data} forms the foundation, serving as the essential input from which models extract and generalize patterns~\cite{wang2023data,data_management4ML_sigmod}. The quality, quantity, and diversity of the data are pivotal in shaping a model's ability to perform effectively across a range of tasks. 
\textit{Algorithms} define how data is processed and guide the architectural design of models, enabling them to address specific problems.
\textit{Computational resources}, such as GPUs, are vital for handling the complex computations required throughout the training process, ensuring scalability and efficiency as models grow in complexity~\cite{angel_ptm_vldb,miao2024demystifying}.

Historically, advancements in AI were primarily driven by model-centric approaches, where the focus was on refining and optimizing algorithmic architectures, such as AlexNet~\cite{NIPS2012_alexnet}, ResNet~\cite{he2016deep}, GoogleNet~\cite{szegedy2015going} in convolutional neural networks. 
However, in recent years, there has been a paradigm shift from the model-centric approach to the data-centric approach for foundation models, such large language models (LLMs). This shift emphasizes the quality, organization, and management of data as the critical factor for improving model performance~\cite{data_management4ML_sigmod,wang2023data,data_collection_vldb}. For example, state-of-the-art LLM models, such as ChatGPT~\cite{openai2022chatgpt}, PaLM~\cite{anil2023palm}, Gemini~\cite{team2023gemini} and Mixtral~\cite{jiang2023mistral} all leverage similar transformer architectures, but they employ distinct data organization and management strategies. 
Therefore, effective training data organization and management has emerged as a decisive factor in scaling model training and enhancing model capabilities~\cite{miao2024demystifying,fernandez2023large,prakash2024integrating,data_collection_vldb}, particularly as models continue to grow in complexity and size.

The long-context capability is one of the most important features in LLMs~\cite{fu2024data,xiong2023effective,li2023loogle}, referring to their ability to process and generate extended text while maintaining coherence, retrieving relevant information~\cite{liu2024lost}, and handling tasks that involve substantial amounts of text efficiently~\cite{bai2023longbench}. For instance, GPT-4~\cite{achiam2023gpt} can handle context windows of up to 128K tokens, and maintain thematic and logical consistency across long-form text. Achieving this long-context capability requires training the model on an extended \textbf{context window}, which usually consist of multiple documents of varying lengths and different domains. 

The primary challenges lie in determining effective data organization and management strategies for long-context model training, which ensure the model achieves long-context capability without compromising its performance in other tasks, such as multitask language understanding~\cite{huang2019unicoder}. Specifically, the strategy of selecting appropriate documents from vast sources of varying lengths to compose each context window during training remains under-explored, yet it is crucial for maximizing model efficiency and performance. 
Existing methods primarily focus on the combination of document sources, aiming to enhance the model's general understanding and generation capabilities. However, they often overlook the consideration of document length, limiting the model's ability to effectively handle long-sequence modeling. 
For example, ICLM~\cite{shi2023context} groups related documents within the same context window, enabling LLMs to access a broader range of relevant contexts, while SPLICE~\cite{staniszewski2023structured} employs retrieval-based strategies~\cite{robertson2009probabilistic,izacard2021unsupervised} to implement structured packing within the context window. 
Unlike previous approaches, we try to explore the data management strategy for long-context model training from a multi-objective combinatorial optimization problem, which involves both document source and length:


\begin{quote}
\em
How to design a data organization and management strategy for LLMs that could achieve long-context capability without compromising the models' performance in other tasks?
\end{quote}

Through extensive experimental analysis, we identified that the challenges associated with data organization and management strategies stem from the following three key aspects:

\textit{\textbf{C1: Shortage of Long Documents Across Multiple Domains.}}
Achieving a balanced domain mixture in training data is essential, as each domain contributes distinct knowledge to the model. However, data from various domains exhibit significant variation in length distribution, complicating the creation of effective long-context training datasets. While some domains provide sufficiently long documents suitable for long-context training, others, particularly web-sourced data, contain shorter documents, with less than 1\% exceeding 4K tokens, as shown in \Cref{fig:domain_stacked_barchart}. To create adequate training contexts from such data, it is often necessary to concatenate multiple documents. This process, however, introduces potential challenges, such as increased cross-attention between unrelated documents, leading to undesirable learning signals and higher computational costs. Consequently, the variation in document length across domains can hinder the model's ability to fully leverage long-context capabilities without introducing noise or redundancy.

\textit{\textbf{C2: Effective Construction for Context Window.}}
Developing robust algorithms for constructing context windows is a critical challenge in enhancing the long-context capabilities of language models. These algorithms must intelligently group related documents to preserve semantic coherence while maximizing the utility of the available data. Poorly designed methods can result in the combination of unrelated or loosely connected documents, which diminishes the quality of training signals and ultimately leads to suboptimal model performance. Although existing research has shown the positive impact of structured data on improving language models' long-context capabilities~\cite{levine2021inductive,chan2022data,devries2023long,devlin2018bert}, these efforts remain in their infancy. Previous approaches~\cite{shi2023context} have frequently resulted in isolated documents, as each document in the training corpus was accessed only once. This limitation necessitates the concatenation of unrelated documents with low semantic correlation in subsequent stages, as illustrated in\Cref{tab:cluster_num_stats} and \Cref{fig:icp_t-SNE}. Moreover, these methods often neglect the essential task of preserving individual document integrity, concentrating on a narrow set of factors rather than adopting a comprehensive framework for context window construction.

\textit{\textbf{C3: Efficient Organization for Large-Scale Datasets.}}
Efficiently organizing large-scale datasets into long-context sequences presents a significant challenge~\cite{big_data_management_vldb,chen2024long}. This task involves not only managing vast amounts of data but also ensuring that the organizational process minimizes computational overhead and labor costs. Inefficient data management strategies can lead to increased training times, elevated computational expenses, and difficulties in effectively scaling the model.
Consequently, optimizing the data organization process is crucial for developing more efficient and scalable large language models~\cite{data_production4ML_sigmod,survey_of_vector_database_management_systems,SW_Store_vldb}. 
Such strategies must ensure a balance between long-context processing and the practical constraints of resource availability, ultimately leading to improved performance and scalability.

To address these challenges, we introduce \textit{\textbf{DataSculpt}}, a novel data management framework designed for long-context training, which formulates the organization of training data as a multi-objective combinatorial optimization (MOCO) problem. 
\textit{DataSculpt} aims to strike a balance among several key objectives, including maximizing relevance and homogeneity among documents within the same context window (\textbf{C1, C2}), preserving document integrity when truncating text that exceeds the context window’s capacity (\textbf{C2}), and optimizing computational efficiency in handling large-scale datasets (\textbf{C3}).
Specifically, our approach employs a coarse-to-fine methodology to optimize the organization of training data efficiently and effectively. 
We start by clustering the data based on semantic similarity (coarse), then perform a multi-objective greedy search within each cluster to score and concatenate documents into different context windows (fine).

To verify the effectiveness of our methods, we conducted continual pretraining with 15 billion tokens on a language model with 7 billion parameters, utilizing varying sequence lengths of 16K, 32K, and 64K. This approach enables the model to continuously adapt and refine its knowledge, resulting in enhanced performance over a one-time pretraining method, making it the most widely used. 
Our experimental results demonstrate that our \textit{DataSculpt} significantly enhances performance across a range of tasks, including general understanding ($+4.88\%$), retrieval augmentation ($+18.09\%$), summarization ($+21.23\%$), reading comprehension ($+21.27\%$), and code completion ($+3.81\%$). 
These findings indicate that \textit{DataSculpt} effectively synergizes the model’s performance in long-context utilization during the continual pretraining phase, even with a limited number of training tokens.

\textit{\textbf{Paper Organization.}}
In \Cref{sec:preliminary}, we introduce the background concepts relevant to this work, including transformer training and the multi-objective combinatorial optimization problem. 
Then, we present a quantitative analysis of the training datasets in \Cref{sec:motivation_and_design}, emphasizing the challenges faced in long-context LLM training.  
In \Cref{sec:method}, we introduce \textit{DataSculpt}, which uses a coarse-to-fine methodology for efficiently organizing training data, enabling the model to achieve long-context capability without compromising performance on other tasks. 
Finally, we validate the effectiveness of \textit{DataSculpt} through comprehensive experiments in \Cref{sec:experiment}.

\section{Preliminary}
\label{sec:preliminary}

\subsection{Transformer Architecture}

Transformers are designed to address sequence modeling and transduction tasks, such as language modeling and machine translation~\cite{vaswani2017attention}. 
At the heart of their architecture lie self-attention networks and point-wise feed-forward networks, which are integral to each layer. 
These models perform dense algebraic operations, including matrix multiplications, which incurs significant computational and memory demands.

\textbf{\textit{Self-Attention Network.}} \quad
Self-attention stands for a critical component of the Transformer architecture, allowing the model to weigh the importance of different words in a sequence, irrespective of their positions~\cite{zhang2019self,Galvatron2023, vaswani2017attention}.  For each word in a sequence, self-attention computes a score that represents how much focus to place on other parts of the input sequence when encoding that specific word. This mechanism involves three sets of weights: query, key, and value, which are learned during training. Mathematically, self-attention can be described as a mapping of a query and a set of key-value pairs to an output, where the output is a weighted sum of the values. Each weight assigned to a value is computed by a compatibility function of the query with the corresponding key. This allows the Transformer to capture complex dependencies and relationships within the data, making it highly effective for tasks that require an understanding of the entire sequence, such as text translation or summarization.


\textbf{\textit{Feed-Forward Network.}} \quad
Each transformer layer includes a position-wise feed-forward network (FFN), which is applied to each position separately and identically. This consists of two linear transformations with a ReLU activation in between. The purpose of the FFN is to transform the representation independently at each position with the same, learned linear transformation. This means that while the self-attention layers share information across positions in the input, the feed-forward network processes each position independently. This component is crucial as it introduces non-linearity into the model, enabling it to learn more complex functions. Moreover, it ensures that the Transformer can adapt to different data types and tasks, providing additional flexibility and power to the model~\cite{vaswani2017attention}.


\subsection{LLMs Training}

Large Language Models (LLMs) denote Transformer-based architectures equipped with hundreds of billions (or more) of parameters~\cite{zhao2023survey}, initially introduced to address sequence modeling and transduction challenges. 
Below, we have outlined several key concepts about LLMs training.

\begin{figure}[t]
    \centering
    \includegraphics[width=0.95\linewidth]{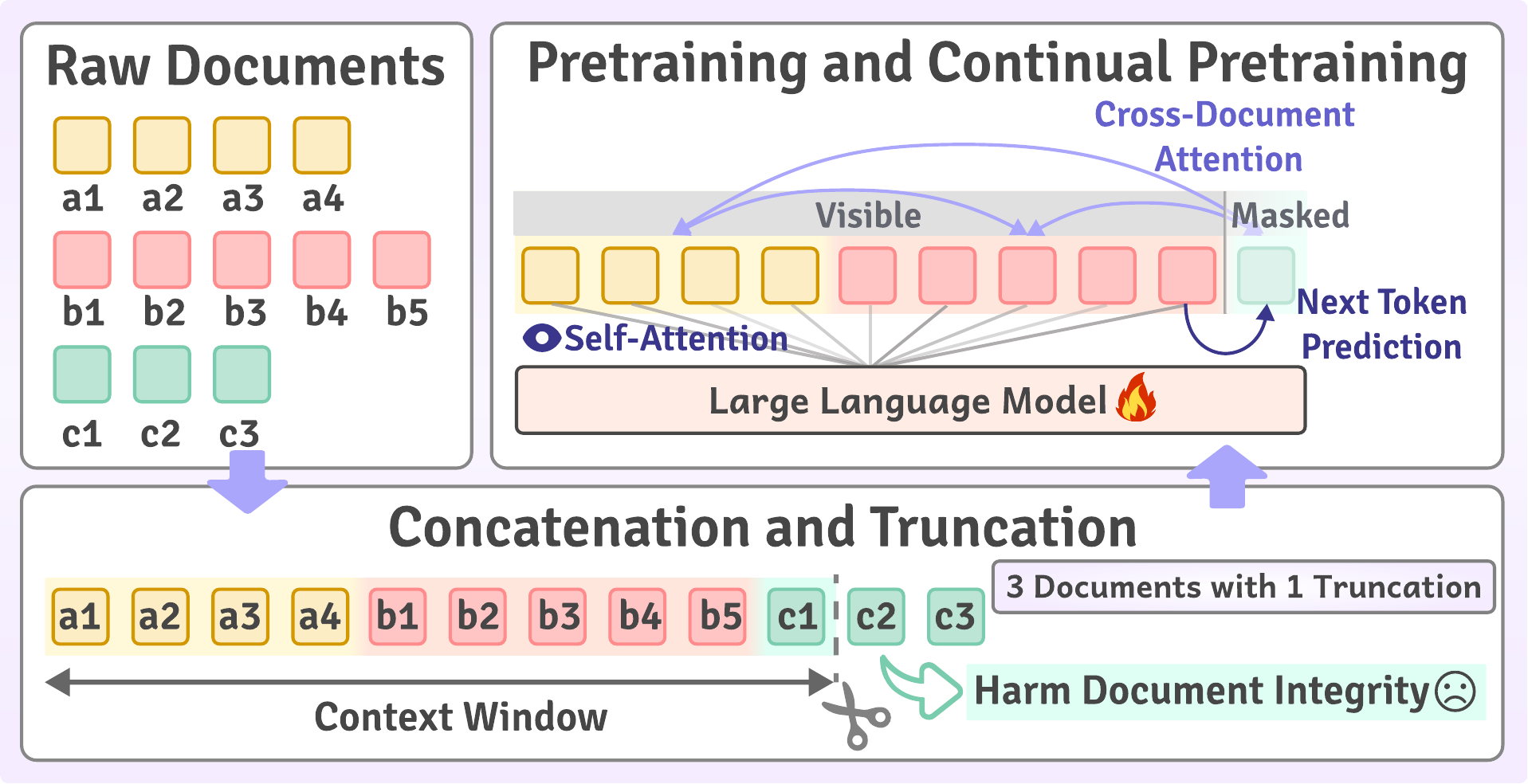}
    \caption{An illustration of the data processing workflow, training task, context window and self-attention mechanism. 
    }
    \label{fig:preliminary}
\vspace{-4mm}
\end{figure}

\textbf{\textit{Pretraining.}} \quad
The pretraining phase of LLMs is a crucial period for developing their general knowledge capabilities. During this phase, LLMs undergo unsupervised training with large-scale raw text to cultivate their language modeling skills. 
Given a sequence $\mathrm {\textbf{x}} = \{x_1, x_2, ..., x_n\}$, the primary training task for LLMs at this stage is to predict subsequent tokens $x_i$ based on preceding tokens $\mathrm {\textbf{x}}_{< i}$ as context, outlined in \cref{fig:preliminary}.
The objective function aims to maximize the likelihood of \Cref{eq:pretraining}.
\begin{gather}
\mathcal{L} _{LLM}^{PT}(\mathrm {\textbf{x}}) = \sum_{i=1}^{n}\log{P(x_i|\mathrm {\textbf{x}}_{< i})}
\label{eq:pretraining}
\end{gather}
\textbf{\textit{Continual Pretraining.}} \quad 
Continual pretraining is a phase that extends the initial pretraining, using the same objective as shown in \Cref{eq:pretraining}. 
The key distinction is that  continual pretraining builds on an already trained model rather than starting from scratch.
At this stage, the LLM has basic language modeling capabilities; the aim now is to enhance its specialized functions, such as \textit{long context modeling}, leveraging the foundational skills acquired during the initial pretraining phase. This phase is increasingly seen as a practical approach to improving model performance without the exorbitant computational costs associated with starting anew, as argued in recent studies~\cite{ibrahim2024simple,gupta2023continual,parmar2024reuse}.

\textbf{\textit{Context Window.}} \quad
The context window is the span of text that a model actively considers when predicting the next word or deciphering the text’s structure. This window defines the range of contextual neighbors used in calculating vector representations~\cite{lison2017redefining}. 
During both the pretraining and continual pretraining phases, raw texts from various documents are \textit{concatenated} to form a continuous sequence that matches the length of the context window. Any text exceeding this length is \textit{truncated}. The size of the context window is pivotal, as it directly impacts the model’s capacity for language comprehension and generation.


\begin{figure*}[t]
    \centering
    \includegraphics[width=\linewidth]{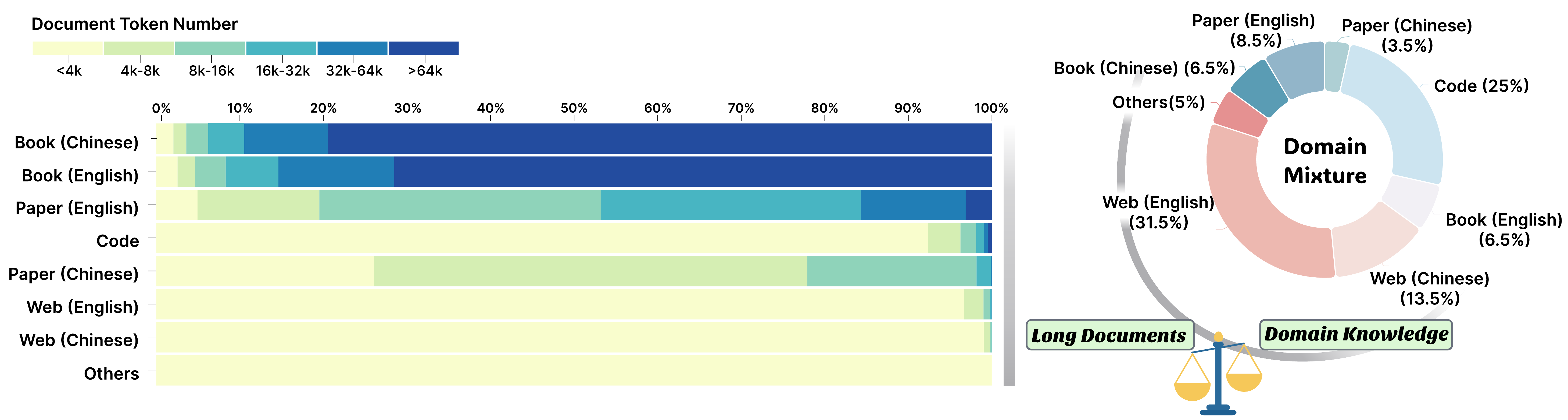}
    \caption{An illustration on the shortage of long documents across multiple domains.
    \textbf{Left}: The length distribution of individual documents across various domains, arranged in descending order by the proportion of documents exceeding 64K tokens. Documents, such as \textit{books} and \textit{academic papers} predominantly conist of medium to long lengths (L $\ge$ 16K), whereas other domains, particularly \textit{web-sourced data}, contain documents almost exclusively shorter than 4K tokens.
    \textbf{Right}: The proportion of each domain in the overall training dataset. }
    \label{fig:domain_stacked_barchart}
\end{figure*}

\begin{table*}[t]
        \centering
        \caption{
        Token statistics across domains. We analyzed the maximum, minimum, and average token counts for different domains. 
        }
        \label{tab:domain_token_stats}
        \begin{tabular}{ccccccccc}
            \toprule
            Domain & \multicolumn{1}{|c}{Book (Chinese)} & Book (English) & Paper (English) & Code & Paper (Chinese) & Web (English) & Web (Chinese) & Others \\
            \midrule
            Max Len & \multicolumn{1}{|c}{24,203,838} & 8,241,838 & 10,721,906 & 1,000,000 & 1,471,297 & 1,267,049 & 80,688 & 21,839 \\
            Min Len & \multicolumn{1}{|c}{667} & 322 & 127 & 173 & 696 & 289 & 132 & 23 \\
            Avg Len & \multicolumn{1}{|c}{176,156} & 216,969 & 20,555 & 2,578 & 6,216 & 886 & 358 & 224 \\
            \bottomrule
        \end{tabular}
\end{table*}

\subsection{\fontsize{10.5}{11.5}\selectfont Multi-Objective Combinatorial Optimization}
\label{subsec:MOCO_pre}

The multi-objective combinatorial optimization (MOCO) problem involves optimizing several objectives simultaneously, which typically conflict with each other~\cite{ulungu1994multi} and can be formulated as:
\begin{align}  
\max_{x \in \chi } \mathbf{f} \left( x\right) & := \left( f_1\left( x\right ), f_2\left( x\right ),...,f_m\left( x\right )\right ) \nonumber \\  
\text{s.t. } & g_k \left(x\right) \le b_k + \epsilon_k, k=1,...,n  
\label{eq:emoco}
\end{align}
where $x$ denotes a solution in the decision space $\chi \subset \mathbb{R}^n$, $\mathbf{f}: \chi \to \mathbb{R}^m$ represents an $m$-dimensional vector-valued objective function, and $g_k \left(x\right) \le b_k + \epsilon_k \, \text{where} \,  k=1,...,n $ stands for a series of soft constraints. Importantly, the evaluation process is computationally expensive for each individual objective $f_i\left(x\right), i = 1, ..., m$. 
The complexity of MOCO challenges lies in the fact that improving the outcome of one objective often leads to the deterioration of another, making it necessary to find a balance or trade-off between them.

\section{Observations and Challenges}
\label{sec:motivation_and_design}





\subsection{Observations on Training Datasets}
The training data is sourced from a variety of domains, such as web content, books, academic papers, and code repositories, with the composition and respective weight proportions shown on the right side of \Cref{fig:domain_stacked_barchart}. 
We conducted a statistical analysis of token count characteristics across various domains, as shown in \Cref{tab:domain_token_stats}. 
Additionally, we categorized tokenized documents into different length intervals based on the token counts of each data source. 
The proportional distributions across domains are visualized as a normalized stacked horizontal bar chart on the left size of \Cref{fig:domain_stacked_barchart}, offering a hierarchical view of the domains according to the prevalence of longer texts.




From \Cref{tab:domain_token_stats} and \Cref{fig:domain_stacked_barchart}, we have observed that data from the \textit{book} and \textit{academic paper} domains predominantly consist of long documents, with over half exceeding 64K tokens in length. 
Academic papers, however, tend to fall within the medium-length range, with documents typically spanning 4K to 32K tokens. In contrast, web-sourced data is primarily composed of short documents, with the vast majority under 4K tokens. As a result, web-sourced data is not naturally suited for long-context training due to its shorter document lengths, which often requires the concatenation of multiple documents to form sequences of adequate length for long-context training.

\begin{figure*}[t]
    \centering
    \includegraphics[width=\textwidth]{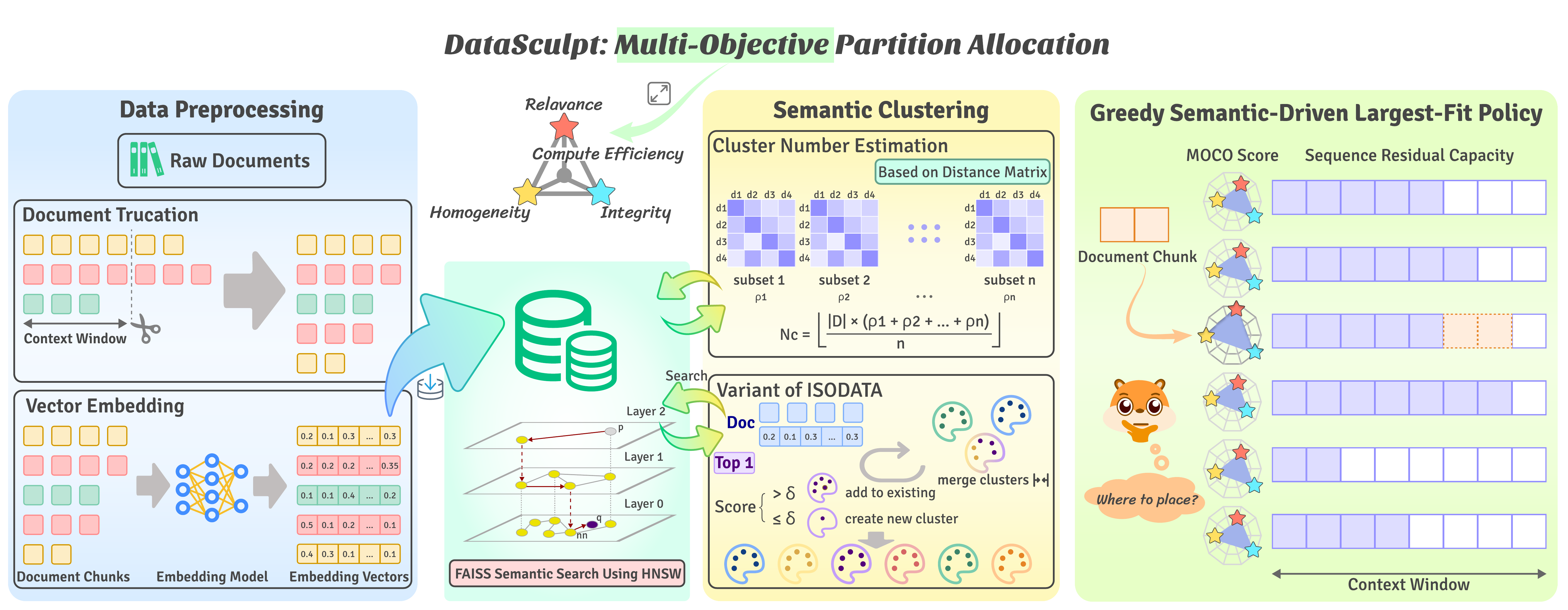}
    \caption{Illustration of DataSculpt. 
    In the data preprocessing stage, documents are initially divided into chunks with $\leq$ context length $L$, which are subsequently transformed into vector embeddings. During the semantic clustering phase, we implement a variant of the ISODATA algorithm, which is augmented with the FAISS vector searching library, to aggregate documents based on semantic similarity. Following this, a greedy semantic-driven largest-fit algorithm is utilized to arrange these clusters into sequences that are multi-objective optimally configured for long-context training.}
    \label{fig:illustration}
\end{figure*}

\subsection{Challenges}

Given the token count statistics across documents, one of the primary challenges lies in the variation of sequence lengths among different data sources.
In efforts to preserve domain diversity, shorter documents in the training dataset tend to increase the incidence of \textit{cross-document attention}, where the model focuses on multiple, unrelated documents within the same context window. 
This can lead to \textit{model hallucinations} and increased \textit{computational consumption}.
The problem becomes more prominent as the model's context window expands during training.


\textbf{\textit{Model Hallucinations.}} 
Model hallucinations occur when models generate incorrect or irrelevant content that is not grounded in the input data. While well-trained models generally reduce cross-document attention, this behavior is not always strictly controlled, which may potentially lead to spurious outputs as well as unintended correlations~\cite{pouransari2024dataset,ding2024fewer}.

\textbf{\textit{Computational Consumption.}} 
Moreover, cross-document attention expends unnecessary computational resources by focusing on unrelated tokens that contribute little to the model's learning process~\cite{krell2021efficient,varivs2021sequence}.  
The unnecessary processing of unrelated tokens not only diminishes the model’s performance but also amplifies the computational demands, which is especially crucial given the attention mechanism's quadratic complexity~\cite{vaswani2017attention}. 

\subsection{Problem Formulation}
\label{subsec:MOCO}

Based on the above observations and analysis, we frame the scenario of data organization and management for long-context training as a \textbf{\textit{multi-objective partition allocation problem with soft constraints}} at a massive data scale. 
This is fundamentally an instance of the multi-objective combinatorial optimization (MOCO) problem~\cite{lin2022pareto} described in \Cref{subsec:MOCO_pre}.

\textbf{Setup.} \quad 
Given a training dataset $\mathcal{D}$ and training sequences $\mathcal{W}$ awaiting sampling, with each context window limited to a length of $L$, the multi-objective partition allocation problem focuses on optimizing several key objectives. The overarching goal is to efficiently enhance the homogeneity of concatenated documents within the same attention sequence:

\begin{itemize}
    \item \textbf{G1: Relevance.} The primary goal is to maximize semantic coherence among documents within the same context window. Increasing relevance has been shown to reduce language model perplexity, improving overall model performance~\cite{shi2023context}.
    \item \textbf{G2: Homogeneity.} Reducing cross-document attention during the model's training phase is crucial. This can be achieved by minimizing the number of concatenated documents within each context window, ensuring the model focuses on fewer, more cohesive sequences.
    \item \textbf{G3: Integrity.}Preserving the integrity of individual documents by minimizing unnecessary truncation during concatenation is essential. This ensures that the contextual richness and completeness of each document are retained as much as possible~\cite{ding2024fewer}.
    \item \textbf{G4: Efficiency.}Efficiency is a key concern when managing large-scale training datasets for LLMs. Achieving a balance between accuracy, processing speed, and space consumption is critical for optimizing long-context training performance while minimizing resource usage.
\end{itemize}


Soft constraints ensure that the total length of documents within each context window does not exceed its capacity, aiming to maximize document integrity and minimize the proportion of truncated tokens within the same document (\textbf{G3}). 
Consequently, \Cref{eq:emoco} can be written as:
\begin{gather}  
\max_{x \in \chi } \mathbf{f} \left( x\right) := \alpha f_1\left( x\right ) +  \beta f_2\left( x\right ) + \lambda p\left(x\right)
\label{eq:emoco2}    
\end{gather}
where soft constraints are incorporated as penalty terms, while $\alpha$, $\beta$, $\lambda$ are non-negative hyper-parameters that control the weight of each objective and penalty function.

The respective objective functions can be formulated as:
\begin{equation}  
\begin{aligned}
f_1\left( x\right ) & = \sum_{w \in \mathcal{W} }\sum_{d_i \in \mathcal{D}}\sum_{\substack{d_j \in \mathcal{D} \\ j\ne i}}x_{d_iw}x_{d_jw}\cos\left(E(d_i),E(d_j)\right) \\  
f_2\left( x\right ) & = \sum_{w \in \mathcal{W} }\frac{L}{\sum_{d_i \in \mathcal{D}}x_{d_iw}}
\end{aligned}
\end{equation}
where $f_1\left( x\right )$ aims to ensure the overall similarity of concatenated documents (\textbf{G1}), while $f_2\left( x\right )$ seeks to guarantee the homogeneity within a training sequence (\textbf{G2}). 
$E(d_i)$ denotes the embedding of document $d_i \in \mathcal{D}$ employed embedding model~\cite{chen2024bge} and cosine similarity is used to determine the similarity between any pair of documents, while $x_{d_iw}$ equals 1 if $d_i \in w$ and 0 otherwise. $L$ is used to divide the objective function $f_2\left( x\right )$ to transform the minimization problem into a maximization problem for resolution.

Furthermore, the penalty term $p(x)$ is defined as:

\begin{equation}
 p(x) = 
    \begin{cases}  
        1,  \text{if } \sum_{w \in \mathcal{W} }\sum_{d_i \in \mathcal{D}}x_{d_iw}l(d_i) \leq L & \\  
        \frac{L}{\sum_{w \in \mathcal{W} }\sum_{d_i \in \mathcal{D}}x_{d_iw}} - 1, \text{otherwise} &
    \end{cases} 
\end{equation}

\noindent where $l(d_i)$ represents the length of document $d_i$. The penalty term ensures that the total length of concatenated documents remains within the capacity $L$ of the context window. If the total sequence length exceeds $L$, the penalty increases to discourage overfitting the context window.

Solving MOCO problems exactly is NP-hard, but coarse-to-fine algorithms are known for providing efficient approximate solutions (\textbf{G4}). The primary challenge, however, lies in effectively balancing multiple objectives within the constraints of limited computational resources and the vast scale of the training corpus $\mathcal{D}$~\cite{zhao2005hierarchical}.

Let $|\mathcal{D}|$ denote the total number of documents in $\mathcal{D}$. The number of documents $k$ selected for concatenation within a given context window is significantly smaller than the total document count, such that $k << |\mathcal{D}|$. As a result, even though score-based sorting is performed on the remaining $|\mathcal{D}| - k$ documents during each iteration, they remain unselected, consuming substantial time and computational resources.

To address this inefficiency, we first apply semantic clustering to the entire training corpus $\mathcal{D}$ at a coarser granularity. This is followed by a greedy algorithm that optimizes the trade-off between computational efficiency and precision during the document organization process.





\section{DataSculpt}
\label{sec:method}

In this work, we introduce \textbf{\textit{DataSculpt}}, a novel data management framework designed for long-context training. This framework aims to effectively integrate data from multiple domains and optimize the context window during training.
\Cref{fig:illustration} presents the workflow of \textit{DataSculpt}. 
In the \textbf{data preprocessing} phase, we first tokenize the raw documents from the training corpus, $\mathcal{D}$, according to the context window length, $L$, of the model.  This process involves dividing the documents into segments, each containing no more than $L$ tokens, to comply with the model's context length constraints. Following this, each document segment $d_i$ is processed through the embedding model to generate its embedding $E(d_i)$.


\subsection{Phase 1: Semantic Clustering}
\label{subsec:semantic_clustering}

\begin{algorithm}[tb]
\caption{Variant of ISODATA}
\label{alg:isodata}
\SetAlgoLined
\SetKwProg{Fn}{Function}{}{end}
\SetKwInput{KwData}{Input}
\SetKwInput{KwParameter}{Parameter}
\SetKwInput{KwDefine}{Define}
\SetKwInput{KwResult}{Ouptut}
\KwData{Training corpus $\mathcal{D}$, Cluster number $N_c$ gained from \Cref{eq:cluster_num}, Hyperparameters: distance threshold $\delta$, alteration threshold $\epsilon$, maximum iterations $T$\;}
\KwParameter{Unclustered data $\mathcal{U}$, Alteration distance $dst$\;}
\KwResult{Clusters $\mathcal{C} = \left\{c_j\right\}_{j=1}^{N_c}$\;}
\KwDefine{$\mu(c)$ is center vector of cluster $c$\;}
Sample $N_c$ centroids in $\mathcal{C} = \left\{c_j\right\}_{j=1}^{N_c}$ randomly from $\mathcal{D}$, $\mu(c_j) \gets E(d_j) \, \text{for all } j = 1, ..., N_c$ \;
$\mathcal{U}, dst \gets [\,],\, 0$ \;
\For{$t = 1, 2, \ldots, T$}{
    \For{each documnet $d_i$ in $\mathcal{D}$}{
        \If{$\max_{c_j \in \mathcal{C}} cos\left(E\left(d_i\right), \mu \left(c_j \right) \right) > \delta$ \textbf{or} $t == T$}{
            Add $d_i$ to $c_{j^*} = \arg\max_{c_j \in \mathcal{C}} cos(E(d_i), \mu(c_j))$\;
        }{
            $\mathcal{U}.append(d_i)$\;
        }
    }
    \For{each $d_u$ in $\mathcal{U}$}{
        Add new cluster $c_u$, $\mu(c_u) \gets E(d_u)$, $N_c \gets N_c + 1$\;
    }
    \For{each cluster $c_j$}{
        $dst \gets dst + || \mu(c_j) -  \frac{1}{|c_j|} \sum_{d_i \in c_j} E(d_i) ||$\;
        Recalculate centroid $\mu(c_j) \gets \frac{1}{|c_j|} \sum_{d_i \in c_j} E(d_i)$\; 
    }
    \For{each pair of clusters $c_i$ and $c_j$}{
        \If{$cos(\mu(c_i), \mu(c_j)) > \delta$}{
            Merge cluster $c_i$ and cluster $c_j$, $N_c \gets N_c - 1$\;
        }
    }
    \If{$dst < \epsilon$}{
        \textbf{Break}\;
    }
    $dst \gets 0$\;
}
\Return Clusters $\mathcal{C} = \left\{c_j\right\}_{j=1}^{N_c}$ \text{where} $\cup_{j=1}^{N_c}c_j=\mathcal{D}$ \;
\end{algorithm}

In the initial phase, the documents undergo a process of semantic clustering at a relatively coarse granularity. To estimate the initial cluster number, we derive inspiration from the construction of a distance matrix by quantifying semantic distances between documents within the training corpus $\mathcal{D}$. 
We first divide $\mathcal{D}$ into $n$ subsets $D_1, D_2, ..., D_n$, where each subset $D_j = \{ d_1, d_2, ..., d_k \} \subset \mathcal{D}$ and $\bigcup_{j=1}^{n}D_j = \mathcal{D}$. 
For each subset $D_j$, we compute the density $\rho_j$ of its distance matrix, which demonstrates the average dissimilarity score in relation to the entire subset. 
\begin{equation}
\label{eq:density}
    \rho_j = \frac{2}{n(n - 1)} \sum_{d_i \in D_j}\sum_{\substack{d_k \in D_j \\ k\ne i}} \cos\left(E(d_i),E(d_k)\right)
\end{equation}

Ultimately, the approximate number of clusters $N_c$ for semantic clustering is ascertained to be equivalent to the total number of documents within the training corpus multiplied by the average score of dissimilarity among subsets $D_1, D_2, ..., D_n$, denoted as
\begin{equation}
\label{eq:cluster_num}
    N_c = \left\lfloor |\mathcal{D}| * \bar{\rho} \right\rfloor \\
        = \left\lfloor |\mathcal{D}| * \frac{\sum_{D_j \subset \mathcal{D}} \rho_j}{n}  \right\rfloor
\end{equation}

We have implemented modifications to the ISODATA clustering algorithm~\cite{memarsadeghi2007fast}, tailoring the algorithm's splitting and merging phases to better align with the specific characteristics of our text corpus, as outlined in \Cref{alg:isodata}. 
Furthermore, to enhance the efficiency of vector similarity assessments, we have integrated the FAISS index~\cite{johnson2019billion}, a sophisticated method for conducting high-speed pairwise similarity searches.

As shown in \Cref{alg:isodata}, we first randomly select $N_c$ data samples from the entire training corpus $\mathcal{D}$ to serve as the initial clustering centers (Line 1), which is based on the calculations from \Cref{eq:cluster_num}. These sampled data points are then indexed using the FAISS vector searching library. Throughout the clustering process with a maximum iteration count of $T$ (Line 3-18), for each document $d_i$, we employ the HNSW method~\cite{malkov2018efficient} from FAISS to retrieve its nearest cluster center $c_j$ (Line 4-7). If the semantic similarity between the current center and $d_i$ exceeds a predefined threshold $\delta$, then $d_i$ is assigned to cluster $c_j$ (Line 5-6). Conversely, if the similarity does not surpass $\delta$, $d_i$ is designated as a new cluster center (Line 7). 
Moreover, as the iterative clustering progresses, adjustments to the cluster centers are made by calculating the mean of the document embedding vectors within each cluster, thereby establishing new cluster centroids (Line 10-12). Additionally, if the similarity between any two clusters exceeds the threshold $\delta$, those clusters are merged (Line 13-15). This approach not only refines the clustering process by dynamically adjusting cluster centers based on semantic coherence but also ensures that clusters remain distinct and meaningful by merging overly similar clusters, thereby enhancing the overall structure and interpretability of the clustering outcome.

\textbf{Time Complexity.} \quad The computational complexity of the original ISODATA algorithm is denoted as $\mathcal{O}(|D|TN_c)$, where $|D|$ represents the quantity of the entire dataset, $N_c$ denotes the initial cluster count, and $T$ signifies the number of iterations. 
However, the integration of FAISS library leveraging HNSW search significantly accelerates the querying for relative vectors through the establishment of a multi-layered indexing architecture, which optimizes the process of pinpointing similar data points across expansive datasets. 
The construction phase of the index is typically completed within a time frame of $O(N_c \log N_c)$, which then supports the rapid search for each query vector. By executing a hierarchical traversal within the index structure, the algorithm efficiently locates the nearest neighbor, showcasing a time complexity of $\mathcal{O}(\log N_c)$. This optimization reduces the overall time complexity to $\mathcal{O}(|D|T\log N_c)$, representing a critical step towards improving the computational efficiency of our data scheduling procedures, enabling more rapid and precise clustering outcomes (\textbf{G4}).

\subsection{Phase 2: Greedy Policy}
\label{subsec:greedy_policy}


In the preliminary phase, documents within the training corpus $\mathcal{D}$ are clustered based on their semantic similarities at a broader level of granularity. 
The subsequent stage leverages a sophisticated multi-objective partition allocation algorithm, strategically applied across these preliminary clusters. The purpose of this algorithm, as depicted in \Cref{alg:greedy_allocation}, is to systematically arrange documents into various context sequences for concatenation. 

\begin{algorithm}[tb]
\caption{Semantic-Driven Largest-Fit Allocation}
\label{alg:greedy_allocation}
\SetAlgoLined
\SetKwProg{Fn}{Function}{}{end}
\SetKwInput{KwData}{Input}
\SetKwInput{KwParameter}{Parameter}
\SetKwInput{KwDefine}{Define}
\SetKwInput{KwResult}{Ouptut}
\KwData{Cluster $c_j \in \mathcal{C}$, Context sequence capacity $L$, Context sequence quantity $N_w$, Hyperparameters: impact of objective and penalty function $\alpha,\,\beta,\,\lambda$\;}
\KwParameter{Available context sequence $\mathcal{A}$, Semantic relevance $f_1$, Residual capacity proportion $f_2$, Truncation penalty $p$, Concatenation score $F$\;}
\KwResult{Constructed sequences $\mathcal{S} = \left\{s_l\right\}_{l=1}^{N_w}$\;}
\KwDefine{$r(s)$ is remaining capacity of context sequence $s$\;}
\KwDefine{$l(d)$ is token quantity of document $d$\;}
\KwDefine{$\mu(s)$ is center vector of context sequence $s$\;}
$\left\{r(s_l) \gets L\right\}_{l=1}^{N_w}$: Initialize empty context sequences\;
$S_{c_j}$: Sort $d_i \in c_j$ by token quantity in descending order\;
\For{each documnet $d_i$ in $S_{c_j}$}{
    $\mathcal{A} = \left\{ s_l | r(s_l) \ge 0\right\}_{l=1}^{N_w}$\;
    \If{$\mathcal{A} = \emptyset$}{
        \textbf{Break}\;
    }
    \For{each context sequence $s_a$ in $\mathcal{A}$}{
        $f_1(d_i, s_a) = cos(E(d_i), \mu(s_a))$\;
        $f_2(d_i, s_a) = r(s_a) / L$\;
        $p(d_i, s_a) = 
            \begin{cases}  
                1, \text{if } l(d_i) \leq r(s_a) & \\  
                L / (L + l(d_i) - r(s_a)), \text{otherwise} & 
            \end{cases} $\;
        $F(d_i, s_a) = \alpha f_1(d_i, s_a) + \beta f_2(d_i, s_a) + \lambda p(d_i, s_a)$\;
        Add $d_i$ to $s_{a^*}$ where $s_{a^*} = \arg\max_{s_a \in \mathcal{S}}F(d_i, s_a)$\;
    }
    $r(s_{a^*}) \gets r(s_{a^*}) - l(d_i)$\;
    $\mu(s_{a^*}) \gets ((|s_{a^*}| - 1)\mu(s_{a^*}) + E(d_i)) / |s_{a^*}|$\;
}
\Return Constructed sequences $\mathcal{S} = \left\{s_l\right\}_{l=1}^{N_w}$\;
\end{algorithm}

As delineated in \Cref{subsec:MOCO}, our objective encompasses augmenting the relevance (\textbf{G1}) and homogeneity (\textbf{G2}), while maintaining integrity (\textbf{G3}) of the textual content encapsulated within each context window. This enhancement is realized through a strategic reduction in the quantity of documents concatenated into a single context sequence, amplifying the semantic congruence among these documents, and eliminating unnecessary truncations.

Initially, documents within each previously formed cluster are organized in descending order by their token count (Line 2). Following this organization, for each document $d_i$ pending allocation, the process seeks to integrate it into an existing sequence $s_a$ with content length that does not exceed the maximum context window capacity $L$ (Line 3-14). The sequence $s_a$ is selected based on achieving the highest composite score $F(d_i, s_a)$, which reflects a weighted consideration of semantic similarity (\textbf{G1}, Line 8), available residual capacity (\textbf{G2}, Line 9), and the minimization of truncation penalties (\textbf{G3}, Line 10).
Through this approach, we aim to ensure that the allocation of documents to sequences is not only computing efficient but also optimally maintains semantic consistency, maximizes capacity utilization, and preserves the integrity of individual documents within each training sequence (Line 11-12). 
The strategy is pivotal in enhancing the overall effectiveness of context sequences for LLMs, by fostering a training environment that is both informationally rich and contextually coherent.

\subsection{Semantic Clustering Analysis}
\label{analytical_study}







\begin{table*}[h]
    \centering
    \caption{Document statistics among clusters. The clusters formed through ICLM's greedy traversal of an incomplete graph demonstrate a significant bias in the distribution of document quantities, with 99.8\% of clusters containing fewer than 100 documents and 53.56\% comprising only a single document. In contrast, clusters generated by DataSculpt's semantic clustering approach display a more balanced distribution of document numbers.}
    \label{tab:cluster_num_stats}
    \begin{tabular}{cccccccc}
        \toprule
        \multicolumn{1}{c|}{\multirow{2}{*}{Method}} & \multirow{2}{*}{Cluster Number} & \multicolumn{6}{|c}{Stats of Docs Number in Clusters} \\
        \cline{3-8}
        \multicolumn{1}{c|}{} & & \multicolumn{1}{|c}{Max} & Min & Mean & Median & Less than 100 & Single Doc \\
        \midrule
        ICLM & \multicolumn{1}{|c}{323,428,858} & \multicolumn{1}{|c}{26,630} & 1 & 3.458 & 1 & 322,926,996 & 173,231,427 \\
        DataSculpt & \multicolumn{1}{|c}{611,981} & \multicolumn{1}{|c}{11,747} & 1 & 1,827 & 763 & 1,479 & 29 \\
        \bottomrule
    \end{tabular}
\end{table*}

\begin{figure*}  
    \centering  
    \begin{subfigure}[b]{0.491\textwidth} 
        \includegraphics[width=\linewidth]{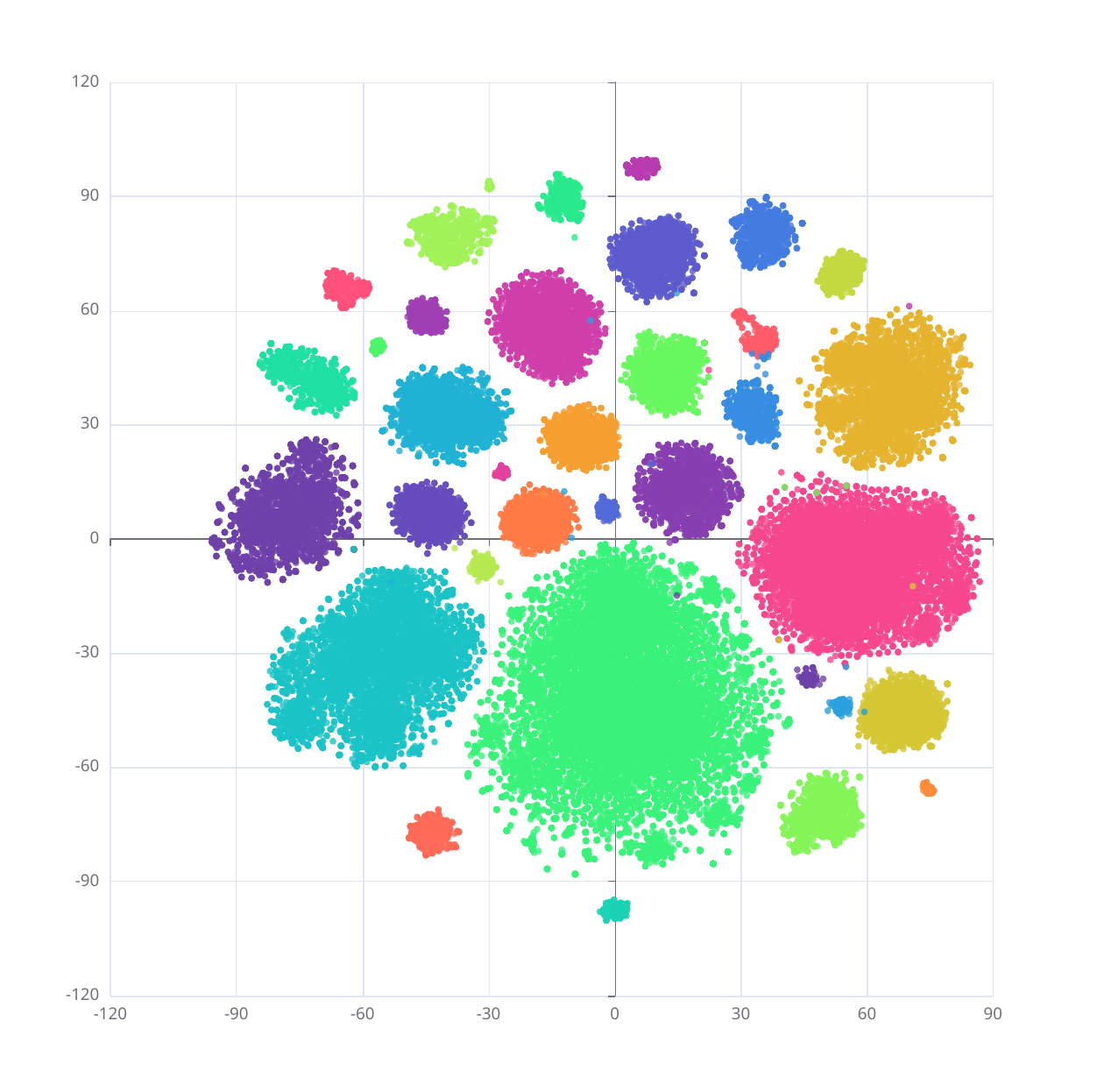}  
        \caption{T-SNE on clusters built from \Cref{alg:isodata}.}  
        \label{fig:t-SNE}  
    \end{subfigure}  
    \hfill 
    \begin{subfigure}[b]{0.491\textwidth} 
        \includegraphics[width=\linewidth]{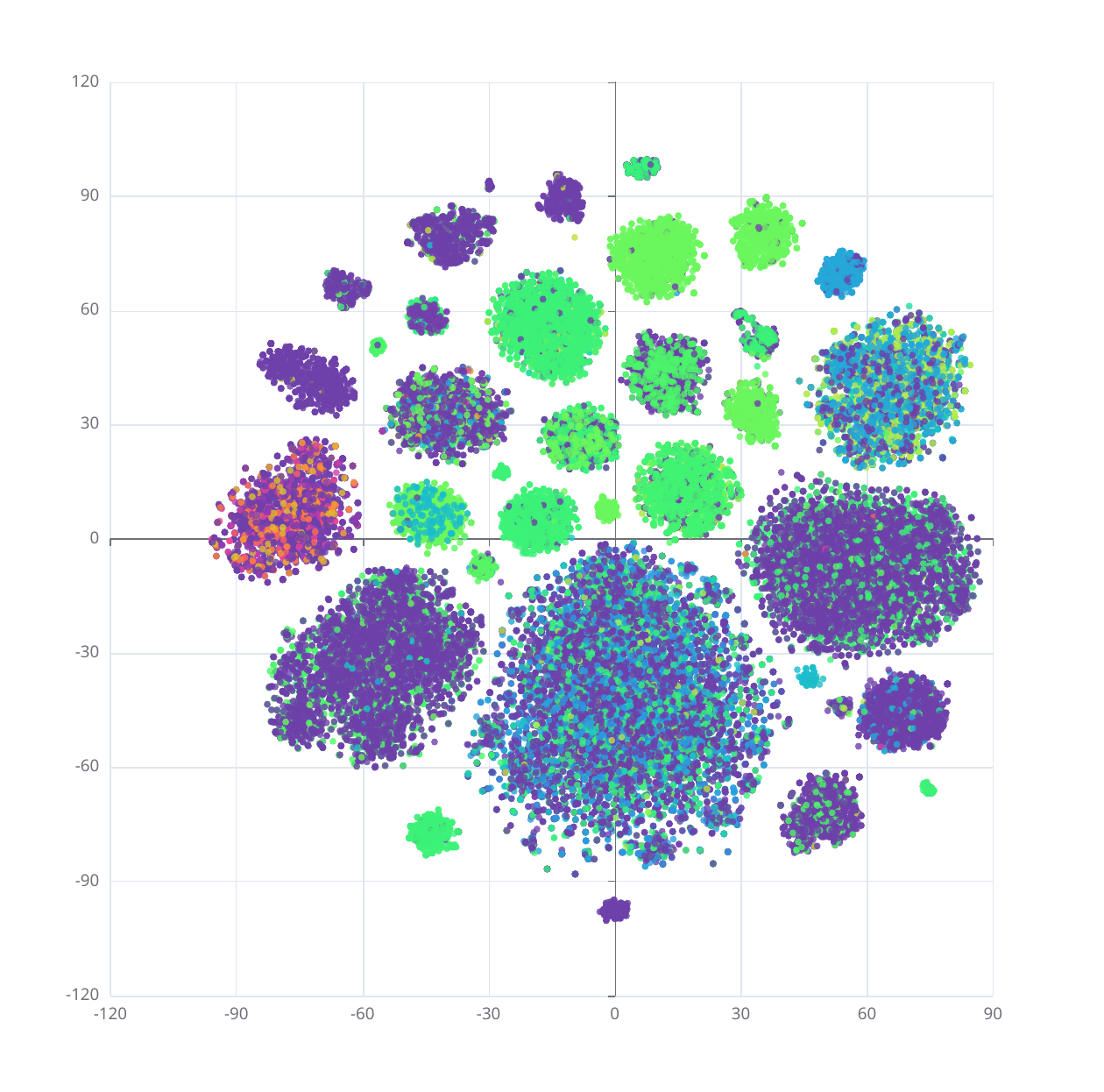}
        \caption{T-SNE on clusters built from graph traversal in ICLM.} 
        \label{fig:icp_t-SNE}  
    \end{subfigure}
    \caption{Visualization using t-SNE on document embeddings sampled from the web derived (English) training corpus, with each cluster denoted by a distinct color. Left~(\ref{fig:t-SNE}) presents the semantic clustering results generated by the methodology outlined in \Cref{alg:isodata}, and right~(\ref{fig:icp_t-SNE}) illustrates clusters where nodes are united by a shared traversal path, as identified by the greedy graph traversal algorithm discussed in ICLM~\cite{shi2023context}.}
    \label{fig:icp_semantic_tSNE}  
\end{figure*}

To demonstrate the robustness of our approach, we compared the effectiveness of semantic clustering across different methodologies.
Our analysis encompasses two key dimensions: the statistical features related to the document quantities within each cluster, and the application of t-SNE~\cite{van2008visualizing} for dimensionality reduction to project and analyze clusters. These clusters were derived from two distinct approaches: the greedy traversal of incomplete graphs as described in ICLM~\cite{shi2023context}, for which we refer to the official implementation~\footnote{https://github.com/swj0419/in-context-pretraining}, and the semantic clustering achieved through an ISODATA variant (\Cref{alg:isodata}) implemented by our DataSculpt.

\subsubsection{Numerical Narratives}
Initially, we performed a quantitative analysis of the number of clusters generated by two distinct methodologies, as detailed in \Cref{tab:cluster_num_stats}. 
It revealed that the clusters produced through greedy graph traversal in ICLM totaled 323,428,858, markedly surpassing the 611,981 clusters formed via semantic clustering with \Cref{alg:isodata}, which suggests a more fragmented clustering approach by the former method. Moreover, clusters initiated at the commencement of the traversal process encompassed a substantially larger number of documents compared to those established later, leading to a pronounced imbalance in document distribution.
Specifically, a significant skew was observed in the distribution of clusters from ICLM, with \textbf{53.56\%} comprising isolated single-document clusters. The dominance of such minimal-scale clusters hinders the ability to adequately fill the context window during the concatenation process for long sequences, thereby inadequately mitigating cross attention in training despite the potential for high semantic similarity among documents. 
Conversely, DataSculpt exhibited a more equitable distribution of cluster sizes throughout the semantic clustering process, enabling the majority of clusters containing semantically akin documents to extend to the length required for a long context sequence. This strategy efficiently capitalizes on the advantages of clustering similar documents, which significantly enhances the model’s training efficiency by promoting semantic coherence within clusters.

\subsubsection{T-SNE Visualization Journeys}
Data visualization is crucial for helping decision-making during massive data management~\cite{data_vis_vldb}. 
Here we employed the dimensionality reduction technique t-SNE to visualize the semantic clusters generated via \Cref{alg:isodata} and those identified through greedy graph traversal on an incomplete graph as described in ICLM. 
For the latter approach, each document was connected to its 20 nearest neighbors, facilitating the construction of the incomplete graph. 

\Cref{fig:icp_semantic_tSNE} clearly demonstrates the effectiveness of \Cref{alg:isodata} in achieving semantic clustering, wherein the documents within each cluster display a pronounced semantic concentration. Upon applying t-SNE for dimensionality reduction and subsequent visualization, distinct clusters can be clearly delineated, showcasing notable semantic boundaries among them. 
This method underscores the capacity of semantic clustering to aggregate documents with high semantic similarity, enhancing the interpretability of large datasets.
However, clusters obtained through the greedy graph traversal in ICLM exhibit substantial size variances, attributed mainly to the restriction of single visitation per document and absence of a strategic path truncation policy.
As a result, such clusters display a wider range of semantic cohesion internally, which leads to a diminished relational coherence among documents concatenated into the same context sequence, potentially undermining the efficacy of the clustering process. 
The ability of \Cref{alg:isodata} to produce semantically concentrated clusters suggests a more robust framework for handling the complexities of large-scale training datasets.

\section{Experiments and Results}
\label{sec:experiment}
In this section, we describe details of our experimental setup (\Cref{experimental_setup}), the baseline methods we use for comparison (\Cref{baselines}), and experimental results (\Cref{results}). 

\subsection{Experimental Setup}
\label{experimental_setup}

We continue pretraining a Transformer decoder-only 7B model~\cite{dong2024baichuanseedsharingpotentialextensive} similar to LLaMA framework~\cite{touvron2023llama} by different sources with the same distribution ratios to the previous pretraining stage, where we structured the data derived from the web (English) domain in accordance with a previously defined multi-objective partition allocation strategy (\Cref{sec:method}). 
The details of training datasets can be found in \Cref{analytical_study}.

\textbf{Data Processing Details.} \quad During the data preprocessing phase, we tokenize raw documents from the training corpus~\cite{baichuan2023baichuan2}, which collectively encompass 855 billion tokens, and then truncated the documents at varying sequence lengths of 16K, 32K, and 64K.
In conducting vector embedding across our training datasets, we employed the BGE-m3 model~\cite{chen2024bge} as our embedding model, encoding each document into a $1024$-dimensional vector. The encoding process necessitated approximately 9 days using a 1,044,090 GPU cluster. 
Following this, we constructed an HNSW index based on FAISS index~\cite{johnson2019billion} to facilitate efficient similarity search for semantic clustering (\Cref{subsec:semantic_clustering}). 
The greedy semantic-driven largest-fit partition algorithm (\Cref{subsec:greedy_policy}) within each cluster was executed on a 4114 CPU Ray cluster.


\textbf{Model Details.} \quad The model comprises a Transformer decoder with 32 attention heads across 32 layers. It features a hidden layer size of 4096, coupled with a dimensionality of 11008 for the forward propagation layers, and the base of Rotary Positional Encoding (RoPE)~\cite{su2024roformer} is modified to accommodate longer context.
We train with 16K, 32K, 64K sequence lengths using 15B tokens for the 7B base model (recall that base model has the 16K context length), applied on 112 H800 GPUs across 14 nodes.
We employ global batch sizes of 224, 112, and 56 sequences per step for context lengths of 16K, 32K, and 64K, which aims at maintaining a uniform number of tokens for model parameter updates.
The learning rate is set to 3e\mbox{-}4 with linear warm-up over the first 1,000 optimization steps and cosine decay schedule. 

\subsection{Baselines}
\label{baselines}

We compare DataSculpt with the following baselines:
(1) \textbf{\textit{Random sampling}} is the standard method that places randomly shuffled and concatenated documents within each input context.
(2) \textbf{\textit{ICLM}}~\cite{shi2023context} utilizes a greedy graph traversal strategy to concatenate semantically similar documents within the context sequence, but only constructing an incomplete graph and restricting each document node to a single visit during traversal due to the extensive volume of the training corpus.

\subsection{Results}
\label{results}

We perform evaluations to validate the effectiveness of DataSculpt on both common capabilities and long-context utilization across various downstream tasks, including general understanding (\Cref{general}), retrieval augmentation (\Cref{retrieval}), summarization (\Cref{summarization}), reading comprehension (\Cref{reading}) and code completion (\Cref{code}).

\subsubsection{General Understanding}
\label{general}

General understanding tasks require the model to comprehend the text holistically, extracting key elements and answering queries, which focuses on broad understanding over specific task execution.
We evaluate the 7B continual pretrained natural language models with zero-shot on MMLU~\cite{hendryckstest2021}, TyDiQA~\cite{clark2020tydi}, BBH~\cite{srivastava2022beyond} and few-shot in-context learning on TREC~\cite{li-roth-2002-learning}.
The evaluation tasks cover different kinds of types, including question answering and text classification. 
Each dataset includes multiple subtasks, and each subtask comes with few-shot examples. 
In \Cref{tab:commonsense}, we compare our models trained via DataSculpt with other baseline models in general understanding across various context window size. 
DataSculpt not only aligns but also slightly outperforms the baselines, particularly in areas requiring multi-hop logical reasoning tasks (+4.88\% compared to random sampling, +4.38\% compared to ICLM).



\begin{table}[h]
    \centering
    \caption{Evaluation on general understanding tasks. Despite DataSculpt targeting to long-context utilization, it achieves performance slightly exceeds that of the baselines.
    }
    \label{tab:commonsense}
    \resizebox{\linewidth}{!}{
    \begin{tabular}{cccccc}
        \toprule
        Context Length & Method & \multicolumn{1}{|c}{MMLU} & TYDIQA & BBH & TREC \\
        \midrule
        \multirow{3}{*}{16K} & random & \multicolumn{1}{|c}{\textbf{52.37}} & 21.45 & 37.87 & 68\\
         & ICLM & \multicolumn{1}{|c}{52.03} & 22.27 & 38.52 & 71 \\
         & DataSculpt & \multicolumn{1}{|c}{51.92} & \textbf{24.75} & \textbf{39.7} & \textbf{71.5}\\
        \midrule
        \multirow{3}{*}{32K} & random & \multicolumn{1}{|c}{47.09} & 17.27 & 33.81 & 70.5 \\
         & ICLM & \multicolumn{1}{|c}{48.09} & 18.8 & 29.95 & 67.5 \\
         & DataSculpt & \multicolumn{1}{|c}{\textbf{48.36}} & \textbf{19.53} & \textbf{35.35} & \textbf{74.5} \\
        \midrule
        \multirow{3}{*}{64K} & random & \multicolumn{1}{|c}{45.26} & 15.28 & 32.9 & 68 \\
         & ICLM & \multicolumn{1}{|c}{\textbf{46.69}} & 16 & 33.14 & 69.5 \\
         & DataSculpt & \multicolumn{1}{|c}{46.27} & \textbf{17.55} & \textbf{34.54} & \textbf{71} \\
        \bottomrule
    \end{tabular}
    }
\end{table}

\begin{table*}[h]
    \centering
    \caption{Key-value retrieval performance on dictionaries of 140 ($\approx$ 11.2K tokens) and 300 ($\approx$ 24K tokens) key-value pairs. $p_i$ represents the scenario where the pertinent information is located at the $i$-th position. “-'' indicates instances where the result is not applicable, as the tokenized context length exceeds the model's capacity. Each position underwent evaluation across 500 examples. The model trained with DataSculpt performs much higher accuracy on hard-to-retrieve positions.}
    \label{tab:kv-retrieval}
    \resizebox{\textwidth}{!}{
        \begin{tabular}{cccccccccccccccc}
            \toprule
            \multirow{2}{*}{Context Length} & \multirow{2}{*}{Method} & \multicolumn{6}{|c}{140 Key-Value Pairs} & \multicolumn{8}{|c}{300 Key-Value Pairs} \\
            \cline{3-8}
            \cline{9-16}
            & & \multicolumn{1}{|c}{$p_1$} & $p_{35}$ & $p_{70}$ & $p_{105}$ & $p_{140}$ & \multicolumn{1}{|c}{Avg.} & \multicolumn{1}{|c}{$p_1$} & $p_{50}$ & $p_{100}$ & $p_{150}$ & $p_{200}$ & $p_{250}$ & $p_{300}$ & \multicolumn{1}{|c}{Avg.} \\
            \midrule
            \multirow{3}{*}{16K} & random & \multicolumn{1}{|c}{97.2} & 59.6 & 20.6 & 16.2 & 8.8 & \multicolumn{1}{|c}{40.48} & \multicolumn{1}{|c}{-} & - & - & - & - & - & - & \multicolumn{1}{|c}{-} \\
             & ICLM & \multicolumn{1}{|c}{\textbf{99.8}} & 20.2 & 25.6 & 13.6 & 11.6 & \multicolumn{1}{|c}{34.16} & \multicolumn{1}{|c}{-} & - & - & - & - & - & - & \multicolumn{1}{|c}{-} \\
             & DataSculpt & \multicolumn{1}{|c}{99} & \textbf{77.6} & \textbf{46.8} & \textbf{36.4} & \textbf{12.6} & \multicolumn{1}{|c}{\textbf{64.48}} & \multicolumn{1}{|c}{-} & - & - & - & - & - & - & \multicolumn{1}{|c}{-} \\
            \midrule
            \multirow{3}{*}{32K} & random & \multicolumn{1}{|c}{82.4} & 48.8 & 46.8 & 38 & 15 & \multicolumn{1}{|c}{46.2} & \multicolumn{1}{|c}{71.8} & 15.8 & 6.4 & 4.4 & 15.2 & 6.6 & 68.6 & \multicolumn{1}{|c}{26.97} \\
             & ICLM & \multicolumn{1}{|c}{68} & 53.8 & 56.2 & 52.6 & 47.2 & \multicolumn{1}{|c}{55.56} & \multicolumn{1}{|c}{75} & 17.7 & 12.3 & 8.6 & \textbf{17.5} & 22.3 & 72.4 & \multicolumn{1}{|c}{32.26} \\
             & DataSculpt & \multicolumn{1}{|c}{\textbf{87}} & \textbf{91} & \textbf{89.4} & \textbf{81.6} & \textbf{51.4} & \multicolumn{1}{|c}{\textbf{80.08}} & \multicolumn{1}{|c}{\textbf{97}} & \textbf{26.2} & \textbf{16} & \textbf{13.6} & 15.6 & \textbf{29.8} & \textbf{83.6} & \multicolumn{1}{|c}{\textbf{40.26}} \\
             \midrule
           \multirow{3}{*}{64K} & random & \multicolumn{1}{|c}{99.6} & 52.4 & 43 & 49.4 & 45.4 & \multicolumn{1}{|c}{57.96} & \multicolumn{1}{|c}{\textbf{99.4}} & 26.6 & 26.2 & 22.4 & 4.2 & 5.4 & 45.4 & \multicolumn{1}{|c}{32.8} \\
             & ICLM & \multicolumn{1}{|c}{92.4} & 74 & 78.8 & 73.6 & 56.8 & \multicolumn{1}{|c}{75.12} & \multicolumn{1}{|c}{88} & 51.8 & 48.2 & 31.4 & 16.7 & 22.6 & 56.3 & \multicolumn{1}{|c}{45} \\
             & DataSculpt & \multicolumn{1}{|c}{\textbf{99.8}} & \textbf{96.1} & \textbf{92.4} & \textbf{87.7} & \textbf{61.5} & \multicolumn{1}{|c}{\textbf{85.5}} & \multicolumn{1}{|c}{97.4} & \textbf{64.4} & \textbf{56.8} & \textbf{38} & \textbf{24.8} & \textbf{30.2} & \textbf{72.6} & \multicolumn{1}{|c}{\textbf{54.89}} \\
            \bottomrule
        \end{tabular}
    }
\end{table*}

\begin{table}[h]
    \caption{Average performance on retrieval augmentation tasks. DataSculpt outperforms the baselines on almost all datasets, with the mean improvement at 19.33\%.}
    \centering
    \label{tab:retrieval1}
    \resizebox{\linewidth}{!}{ 
        \begin{tabular}{ccccc}
            \toprule
            Context Length & Method & \multicolumn{1}{|c}{K-V Retrieval} & TriviaQA & Needle. \\
            \midrule
            \multirow{3}{*}{16K} & random & \multicolumn{1}{|c}{-} & 86.15 & 32.1 \\
             & ICLM & \multicolumn{1}{|c}{-} & 87.86 & 33.7 \\
             & DataSculpt & \multicolumn{1}{|c}{-} & \textbf{89.58} & \textbf{44.9} \\
            \midrule
            \multirow{3}{*}{32K} & random & \multicolumn{1}{|c}{36.59} & \textbf{86.31}  & 49.4 \\
             & ICLM & \multicolumn{1}{|c}{43.91} & 85.48  & 58.2 \\
             & DataSculpt & \multicolumn{1}{|c}{\textbf{60.17}} & 84.12 & \textbf{63.7} \\
            \midrule
            \multirow{3}{*}{64K} & random & \multicolumn{1}{|c}{45.38} & 85.85 & 77.4 \\
             & ICLM & \multicolumn{1}{|c}{60.06} & 86.02 & 82.8 \\
             & DataSculpt & \multicolumn{1}{|c}{\textbf{70.20}} & \textbf{86.84} & \textbf{90} \\
            \bottomrule
        \end{tabular}
    }
\end{table}

\begin{figure*}[h]
    \centering
    \includegraphics[width=\textwidth]{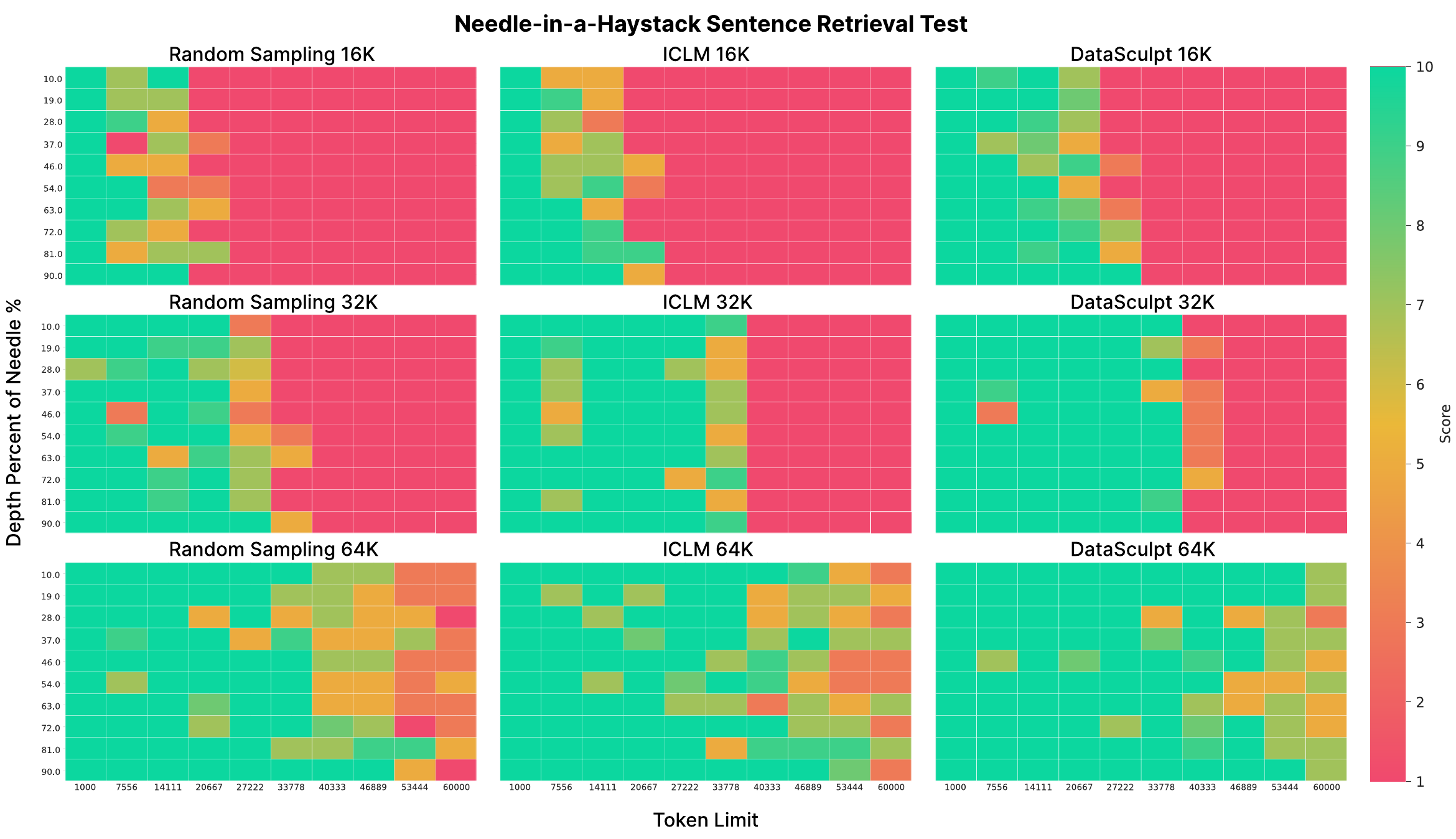}
    \caption{“Needle in a haystack'' performance comparison, where the x-axis represents the document’s length (the “haystack''), and the y-axis reflects the position of the “needle'' (a short sentence) within the document, ranging from 1K to 64K tokens. The model’s performance in reciting information from the “needle'' across various document lengths and positions is assessed on a scale from 1 to 10, represented through a color gradient transitioning from red (score of 1) to green (score of 10). }
    \label{fig:needle_haystack}
\end{figure*}

\subsubsection{Retrieval Augmentation}
\label{retrieval}

Retrieval augmentation involves LLMs to access external knowledge sources, such as databases or text corpora to retrieve pertinent information as required, effectively
extending their memory and knowledge capabilities beyond what is available in their pretrained state. 
In assessing the performance of models on retrieval augmentation tasks, we focus on the models' synthetic key-value retrieval performance~\cite{liu2024lost} under settings of 140 and 300 key-value pairs and conduct a zero-shot evaluation using the TriviaQA dataset~\cite{joshi2017triviaqa}. Further experiments are also undertaken on the “needle in a haystack'' task~\cite{kamradt2023needle,fu2024data} with a maximum of 64K tokens to gain a comprehensive understanding of the effectiveness of these models in identifying specific information within extensive contexts. 

Results in \Cref{tab:kv-retrieval} shows the superior performance of DataSculpt in key-value retrieval tasks across both 140 and 300 key-value pairs configurations, showcasing improvements of up to 59.09\% over random sampling and 34.33\% over ICLM. 
Since it proves challenging for a model trained with 16K sequence length to extrapolate its capabilities to 24K, models trained under such setting did not yield measurable scores in the 300 key-value pairs task.
Additionally, \Cref{fig:needle_haystack} illustrates the visualization outcomes for the “needle in a haystack'' task, indicating that DataSculpt notably enhances the model's ability to extrapolate beyond the training context lengths.

\subsubsection{Summarization}
\label{summarization}

Summarization tasks involve generation of concise and coherent abstracts from a given source text, which serves as a measure of a model's ability to distill key information, maintain the original text's intent, and present it in a brief and comprehensible manner.
Compared to traditional QA and text classification tasks, which predominantly depend on local context for deriving solutions, summarization requires a broader understanding of the entire document~\cite{bai2023longbench}. 
This capability is more closely associated with the long-context processing capabilities of LLMs.

We assess the continued pretrained models in a zero-shot setting on real-world benchmarks including GovReport~\cite{huang2021efficient} as well as QMSum~\cite{zhong2021qmsum}, and employ few-shot in-context learning on SAMSum~\cite{gliwa2019samsum}.
Here we utilize the ROUGE-L~\cite{lin2004rouge} metric, an N-gram based measure extensively adopted in QA and summarization tasks. 
Improvements in summarization tasks achieved through DataSculpt are documented in \Cref{tab:summarization}, showcasing enhancements of 21.21\% over random sampling and 18.18\% over ICLM. 



\begin{table}[h]
    \centering
    \caption{Evaluation on summarization tasks. DataSculpt consistently surpasses baseline models across nearly all three datasets and various context lengths. }
    \label{tab:summarization}
    \resizebox{\linewidth}{!}{
        \begin{tabular}{ccccc}
            \toprule
            Context Length & Method & \multicolumn{1}{|c}{GovReport} & QMSum & SAMSum \\
            \midrule
            \multirow{3}{*}{16K} & random & \multicolumn{1}{|c}{24.51} & 18.6 & 15.83\\
             & ICLM & \multicolumn{1}{|c}{25.5} & 17.43 & 16.67 \\
             & DataSculpt & \multicolumn{1}{|c}{\textbf{27.66}} & \textbf{21.41} & \textbf{20.41}\\
            \midrule
            \multirow{3}{*}{32K} & random & \multicolumn{1}{|c}{24.04} & 18.42 & 12.23 \\
             & ICLM & \multicolumn{1}{|c}{27.71} & \textbf{18.72} & 12.85 \\
             & DataSculpt & \multicolumn{1}{|c}{\textbf{28.83}} & 17.87 & \textbf{24.04} \\
            \midrule
            \multirow{3}{*}{64K} & random & \multicolumn{1}{|c}{26.75} & 18.92 & 12.47 \\
             & ICLM & \multicolumn{1}{|c}{25.92} & 17.82 & 13.65 \\
             & DataSculpt & \multicolumn{1}{|c}{\textbf{29.42}} & \textbf{20.25} & \textbf{18.41} \\
            \bottomrule
        \end{tabular}
    }
\end{table}

\subsubsection{Reading Comprehension}
\label{reading}

Reading comprehension assesses the capacity of models to generate responses to queries, grounded in the context provided within a given textual passage.
We evaluate the models’ ability in reading comprehension with zero-shot on NarrativeQA~\cite{kovcisky2018narrativeqa}, Qasper~\cite{dasigi2021dataset}, HotPotQA~\cite{yang2018hotpotqa} and 2WikiMultihopQA~\cite{ho2020constructing}.
NarrativeQA and Qasper fall into the category of single-document question-answering (QA) datasets, whereas HotPotQA and 2WikiMultihopQA are recognized as Wikipedia-based, multi-hop QA datasets. 
Evaluation across these datasets is conducted using the F1 metric, 
and enhancements attributed to our DataSculpt in reading comprehension tasks are further detailed in \Cref{tab:reading1} and \Cref{tab:reading2}.


\begin{table}[h]
    \centering
    \caption{
    In zero-shot evaluations on the validation subsets of NarrativeQA and Qasper, DataSculpt records improvements of 27.06\% over random sampling and 13.22\% over ICLM. }
    \label{tab:reading1}
    \begin{tabular}{cccc}
        \toprule
        Context Length & Method & \multicolumn{1}{|c}{NarrativeQA} & Qasper \\
        \midrule
        \multirow{3}{*}{16K} & random & \multicolumn{1}{|c}{7.46} & 7.53 \\
         & ICLM & \multicolumn{1}{|c}{12.34} & 9.6 \\
         & DataSculpt & \multicolumn{1}{|c}{\textbf{19.33}} & \textbf{9.83} \\
        \midrule
        \multirow{3}{*}{32K} & random & \multicolumn{1}{|c}{11.09} & 10.55 \\
         & ICLM & \multicolumn{1}{|c}{16.47} & 9.63 \\
         & DataSculpt & \multicolumn{1}{|c}{\textbf{18.33}} & \textbf{11.35} \\
        \midrule
        \multirow{3}{*}{64K} & random & \multicolumn{1}{|c}{16.53} & 9.16 \\
         & ICLM & \multicolumn{1}{|c}{14.07} & \textbf{9.97} \\
         & DataSculpt & \multicolumn{1}{|c}{\textbf{18.45}} & 9.07 \\
        \bottomrule
    \end{tabular}
\end{table}

\begin{table}[h]
    \centering
    \caption{Zero-shot F1 performance on the reading comprehension tasks of HotPotQA and 2WikiMultihopQA datasets, with an average improvement of 10.54\% on HotPotQA and 6.86\% on 2WikiMultihopQA.}
    \label{tab:reading2}
    \resizebox{\linewidth}{!}{
        \begin{tabular}{cccc}
            \toprule
            Context Length & Method & \multicolumn{1}{|c}{HotPotQA} & 2WikiMultihopQA \\
            \midrule
            \multirow{3}{*}{16K} & random & \multicolumn{1}{|c}{3.52} & 8.66 \\
             & ICLM & \multicolumn{1}{|c}{\textbf{8.07}} & 9.57 \\
             & DataSculpt & \multicolumn{1}{|c}{7.68} & \textbf{9.82} \\
            \midrule
            \multirow{3}{*}{32K} & random & \multicolumn{1}{|c}{\textbf{10.09}} & 10.45 \\
             & ICLM & \multicolumn{1}{|c}{8.73} & 10.39 \\
             & DataSculpt & \multicolumn{1}{|c}{8.31} & \textbf{11.24} \\
            \midrule
            \multirow{3}{*}{64K} & random & \multicolumn{1}{|c}{8} & 9.98 \\
             & ICLM & \multicolumn{1}{|c}{8.47} & 9.51 \\
             & DataSculpt & \multicolumn{1}{|c}{\textbf{9.92}} & \textbf{10.23} \\
            \bottomrule
        \end{tabular}
    }
\end{table}

\begin{figure*}[h]
    \centering
    \includegraphics[width=\textwidth]{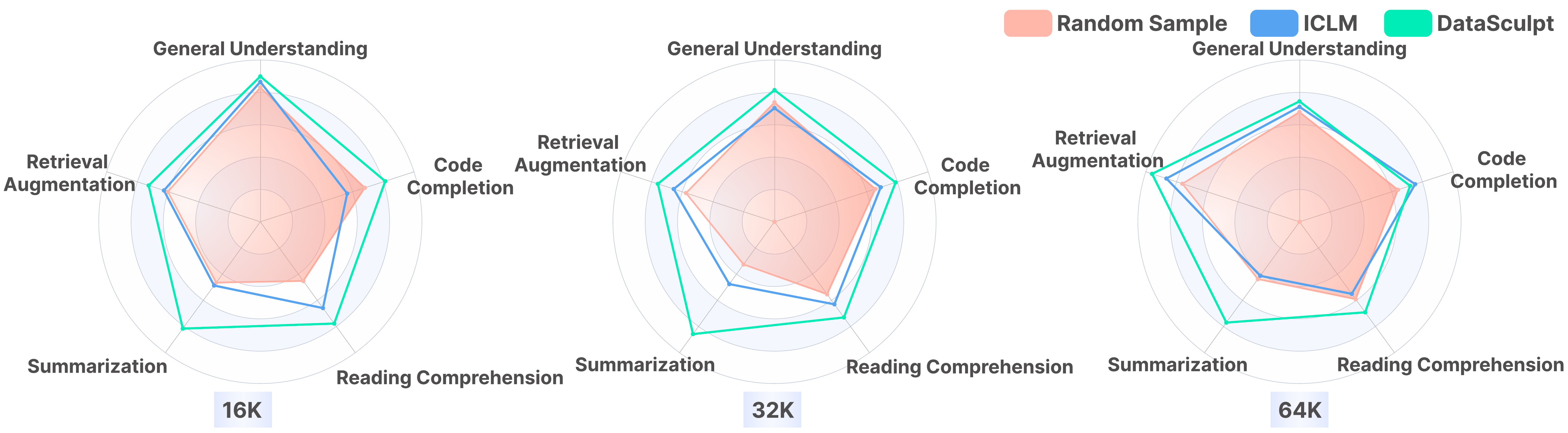}
    \caption{Radar charts depicting the average performance metrics of models trained with varying sequence lengths on a spectrum of benchmark categories. Among all evaluated training lengths, models developed using the DataSculpt approach surpass baseline-trained models in performance metrics across a diverse range of benchmark categories.}
    \label{fig:radar_chart}
\end{figure*}

\subsubsection{Code Completion}
\label{code}

Code completion serves as a critical and challenging task for evaluating a model’s long context modeling ability, which facilitates user coding by predicting and completing code segments based on previous inputs and contextual understanding. 
For code completion tasks, we assess the models using zero-shot learning on the HumanEval dataset~\cite{chen2021evaluating}, LCC~\cite{guo2023longcoder}, and RepoBench-P~\cite{liu2023repobench}, where the latter two datasets are specifically designed for long code modeling evaluation. 
Here we applied execution-based accuracy (Pass@1) for HumanEval evaluation and Edit Sim (Levenshtein distance) for the remaining two code generation tasks. 


\begin{table}[h]
    \centering
    \caption{Performance on code completion, with an enhancement of 4.23\% over random sampling and 3.46\% on ICLM.}
    \label{tab:code_completion}
    \resizebox{\linewidth}{!}{
        \begin{tabular}{ccccc}
            \toprule
            Context Length & Method & \multicolumn{1}{|c}{HumanEval} & LCC & RepoBench-P \\
            \midrule
            \multirow{3}{*}{16K} & random & \multicolumn{1}{|c}{10.37} & 65 & 60.24  \\
             & ICLM & \multicolumn{1}{|c}{10.58} & 64.22 & 55.64 \\
             & DataSculpt & \multicolumn{1}{|c}{\textbf{11.74}} & \textbf{68.28} & \textbf{61.54} \\
            \midrule
            \multirow{3}{*}{32K} & random & \multicolumn{1}{|c}{12.2} & 64.52 & 57.85 \\
             & ICLM & \multicolumn{1}{|c}{12.26} & 62.91 & \textbf{60.94} \\
             & DataSculpt & \multicolumn{1}{|c}{\textbf{12.68}} & \textbf{68.58} & 59.21 \\
            \midrule
            \multirow{3}{*}{64K} & random & \multicolumn{1}{|c}{12.8} & 64.01 & 57.06 \\
             & ICLM & \multicolumn{1}{|c}{12.88} & \textbf{67.11} & 58.9 \\
             & DataSculpt & \multicolumn{1}{|c}{\textbf{13.01}} & 64.37 & \textbf{60.04} \\
            \bottomrule
        \end{tabular}
    }
\end{table}


Results of zero-shot evaluation are presented in \Cref{tab:code_completion}, which reveals a modest advantage of DataSculpt in synergizing average scores for models' code capabilities. While the data was not explicitly structured for the coding domain, advancements in long-context processing achieved through DataSculpt have correspondingly manifested in enhanced coding performance. 

\section{Related Work}
\label{sec:related_work}

\subsection{Data-Centric Long-Context Training}

Training Large Language Models (LLMs) with expanded context windows facilitates the execution of tasks that surpass the capabilities defined by the existing paradigm, such as multi-document question answering~\cite{caciularu2023peek}, long document summarization~\cite{bai2023longbench}, and enhanced retrieval functionalities~\cite{liu2024lost}. 
However, LLMs' efficacy in handling long context does not consistently meet the expected standards. 
Data-focused approaches hold substantial importance~\cite{wang2023data} for enhancing LLMs' long-context capabilities. 

The construction of long-context data involves more than a mere concatenation of short documents, which can lead to cross-document attention within a sequence. 
In these circumstances, formulating high-quality or domain-specific long sequences holds significant research value~\cite{devries2023long}. 
There have been data-centric efforts made in improving LLMs' capabilities for long context processing. 
ICLM~\cite{shi2023context} and SPLICE~\cite{staniszewski2023structured} combine similar training instances through a sophisticated retrieval process, which emphasizes the critical role of high-quality long-context training data in enhancing the context expansion capabilities of LLMs without necessitating extensive data resources.
Additionally, Krell et al.~\cite{krell2021efficient} deliberate on the challenges associated with the conventional concat-and-chunk approach, proposing an approximate bin-packing algorithm as a potential solution. As an improvement, dataset decomposition~\cite{pouransari2024dataset} and Best-fit Packing~\cite{ding2024fewer} both propose the sequence length aware optimization technique to minimize truncations, thereby facilitating more efficient pretraining on long sequences and achieving scalability relative to the size of dataset.
Moreover, ProLong~\cite{chen2024long} introduces an evaluation framework for assessing the quality of long-context data, further enriching the field's methodological toolkit.
Nonetheless, these methods have not performed a systematic framework tailored for data construction in long context training.
Our work conceptualizes data management for extended-context training as a multi-objective combinatorial optimization problem, and provide a comprehensive framework for long sequence data organization.

\subsection{MOCO-Augmented LLMs}


Existing works have integrated Multi-objective combinatorial optimization (MOCO) problems with relevant concerns in the LLMs field.
Wang et al.~\cite{wang2024interpretable} employed the MOCO approach within the Reinforcement Learning from Human Feedback (RLHF) framework for LLMs, where the reward models aggregate multiple reward objectives with coefficients into a scalar score. 
Akiba et al.~\cite{akiba2024evolutionary} viewed model merging as a MOCO challenge, and introduced an innovative approach that leverages evolutionary algorithms to facilitate the automated development of robust foundation models.
Furthermore, integration of MOCO with data-centric LLMs have been explored for training data selection~\cite{du2023mods,xu2023rethinking,chen2024long}, which applies heuristic techniques to assign weights to different data evaluation metrics and facilitates a phased resolution approach.
Considering the extensive scale of training dataset, data construction for long context is treated as a MOCO challenge, for which the coarse-to-fine algorithm is identified as a suitable resolution strategy.

\section{Conclusion}
\label{sec:conclusion}
The ability to process long contexts is crucial for Large Language Models (LLMs), making the effective training of LLMs for extended-context tasks from a data-centric perspective a worthy area of investigation. A pressing challenge lies in organizing long-sequence data for domains lacking extensive text, while ensuring diversity in domain knowledge. In this work, we introduced \textbf{\textit{DataSculpt}}, a framework for organizing long-context training data, conceptualizing the task as a multi-objective combinatorial optimization problem. 
This framework considered factors including relevance, homogeneity, integrity, and computational costs associated with data construction, employing a coarse-to-fine method for phased greedy solutions. 
Experimental results indicated that data organized via \textit{DataSculpt} significantly enhances long-context capabilities without compromising general inference abilities, marking a substantial improvement over existing data management methods.

\clearpage

\bibliographystyle{ACM-Reference-Format}
\bibliography{vldb2025}

\end{document}